\documentclass[letterpaper, 10pt, journal,twoside]{ieeeconf}
\IEEEoverridecommandlockouts                              

\usepackage[english]{babel}
\usepackage{amsmath}
\usepackage{amssymb}
\usepackage{booktabs}
\usepackage{siunitx}
\usepackage[dvips]{graphicx}
\usepackage{multirow}
\usepackage{amsfonts}
\usepackage{enumerate}
\usepackage{tabularx}
\usepackage{algorithm,algorithmic}
\usepackage{bm}
\usepackage{enumerate}
\usepackage{adjustbox}
\usepackage{xcolor}
\usepackage{siunitx}
\usepackage{booktabs}
\usepackage{mdframed}
\usepackage{adjustbox}
\usepackage{authblk}
\usepackage{color}
\usepackage{xurl}
\usepackage{subcaption}
\usepackage{cite}
\makeatletter
\let\NAT@parse\undefined
\makeatother
\usepackage{hyperref}
\usepackage{relsize}
\usepackage{float}
\usepackage{pifont}
\usepackage{tikz}
\usetikzlibrary{positioning,fit}
\usepackage{diagbox}

\usepackage{cuted}
\usepackage{capt-of}
\usepackage{comment}
\usepackage[most]{tcolorbox}

\title{\LARGE \bf NeuroMesh: A Unified Neural Inference Framework \\for Decentralized Multi-Robot Collaboration}

\author{Yang Zhou$^{1}$, Yash Shetye$^{1}$, Long Quang$^{1,2}$, Devon Super$^{3}$, Jesse Milzman$^{2}$, Manohari Goarin$^{1}$,
Aditya Azad$^{1}$, Devang Sunil Dhake$^{1}$, Jeffery Mao$^{1}$,
Carlos Nieto-Granda$^{2}$, and Giuseppe Loianno$^{4}$
\thanks{Manuscript received: December 23, 2025; Revised March 19, 2026; Accepted Apr, 14, 2026. This paper was recommended for publication by Editor M. Ani Hsieh upon evaluation of the Associate Editor and Reviewers' comments. This work was supported by the DEVCOM ARL grant SARA W911NF-24-2-0057, the DARPA YFA Grant D22AP00156-00, and the NSF CPS Grant CNS-2603416.}
\thanks{\textsuperscript{1}The authors are with New York University, New York, USA. {\tt\footnotesize email: \{yangzhou, ys5153, lq2146, mg7363, aa10878, dsd9855, jm7752\}@nyu.edu}.}
\thanks{\textsuperscript{2}The authors are with U.S. Army Combat Capabilities Development Command, Army Research Laboratory, Adelphi, MD 20783, USA.
{\tt\footnotesize email: \{long.p.quang.civ, jesse.m.milzman.civ, carlos.p.nieto2.civ\}@army.mil}.}
\thanks{\textsuperscript{3}The author is with Vanderbilt University, Tennessee, USA. {\tt\footnotesize email: devon.a.super@vanderbilt.edu}.}
\thanks{\textsuperscript{4}The author is with the University of California Berkeley, Department of Electrical Engineering and Computer Sciences, Berkeley, CA 94720, USA. {\tt\footnotesize email: loiannog@eecs.berkeley.edu}.}
\thanks{Digital Object Identifier (DOI): see top of this page.}

}
\begin{document}

\thispagestyle{empty}
\pagestyle{empty}
\makeatletter
\g@addto@macro\@maketitle{
    \setcounter{figure}{0}
    \centering
    \setlength{\fboxrule}{1pt}
    \setlength{\fboxsep}{2pt}
    \def\teasersidewidth{0.115\textwidth}
    \def\teasermiddlewidth{0.55\textwidth}
    \def\teasercolgap{0.01\textwidth}
    \noindent
    \hspace*{\fill}%
    \begin{minipage}[b]{\teasersidewidth}
        \centering
        \fcolorbox{red}{white}{\includegraphics[width=0.88\linewidth]{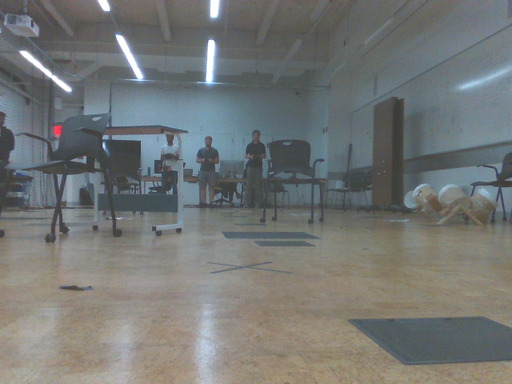}}

        \vspace{0.15em}
        \fcolorbox{green}{white}{\includegraphics[width=0.88\linewidth]{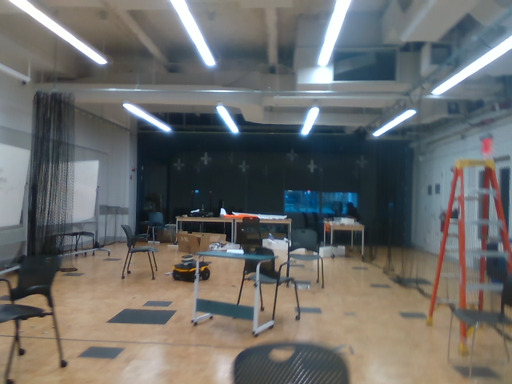}}

        \vspace{0.15em}
        \fcolorbox{blue}{white}{\includegraphics[width=0.88\linewidth]{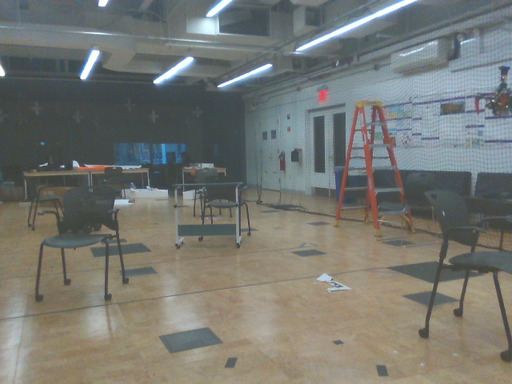}}
    \end{minipage}%
    \hspace{\teasercolgap}%
    \begin{minipage}[b]{\teasermiddlewidth}
        \centering
        \includegraphics[width=\linewidth,trim=0 40 0 0, clip]{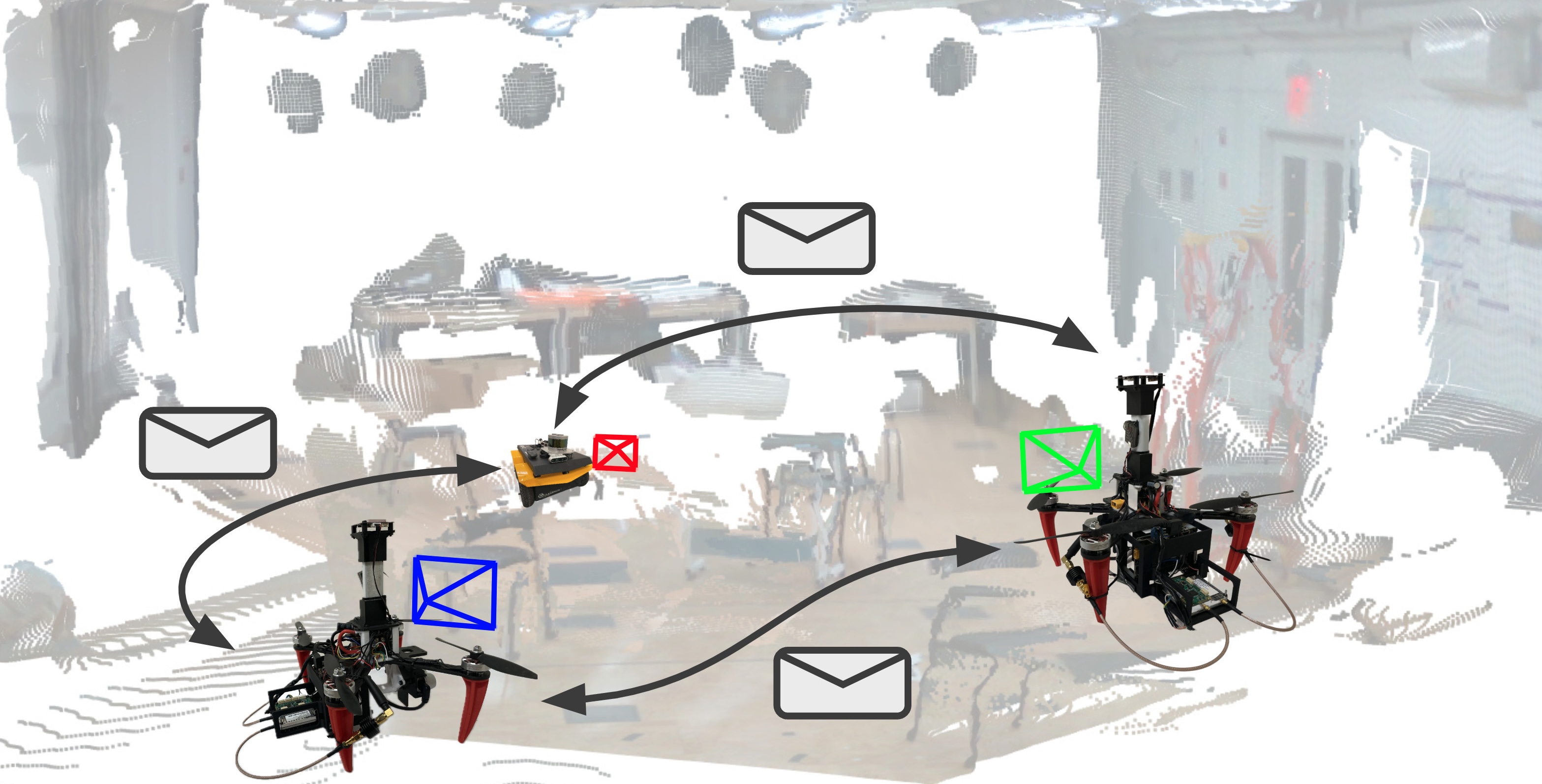}
    \end{minipage}%
    \hspace{\teasercolgap}%
    \begin{minipage}[b]{\teasersidewidth}
        \centering
        \fcolorbox{red}{white}{\includegraphics[width=0.88\linewidth]{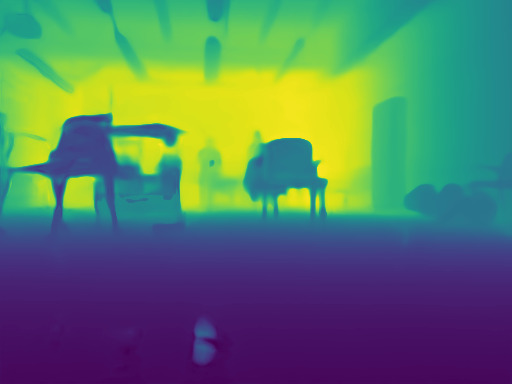}}

        \vspace{0.15em}
        \fcolorbox{green}{white}{\includegraphics[width=0.88\linewidth]{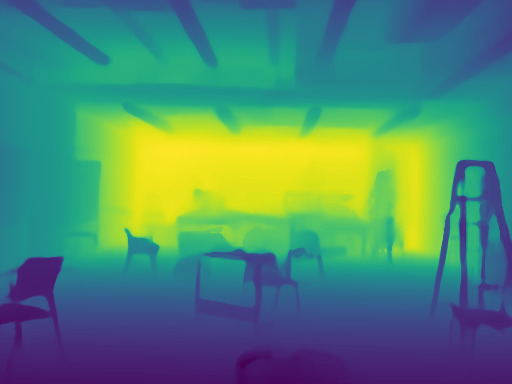}}

        \vspace{0.15em}
        \fcolorbox{blue}{white}{\includegraphics[width=0.88\linewidth]{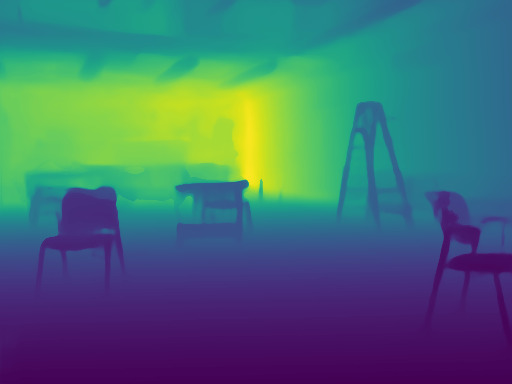}}
    \end{minipage}%
    \hspace*{\fill}

    \vspace{0.4ex}

    {\footnotesize
    \noindent
    \hspace*{\fill}%
    \begin{minipage}[t]{\teasersidewidth}
        \centering
        RGB Input
    \end{minipage}%
    \hspace{\teasercolgap}%
    \begin{minipage}[t]{\teasermiddlewidth}
        \centering
        Reconstructed consistent 3D Pointcloud output by NeuroMesh framework
    \end{minipage}%
    \hspace{\teasercolgap}%
    \begin{minipage}[t]{\teasersidewidth}
        \centering
        Depth Output
    \end{minipage}%
    \hspace*{\fill}}
    \vspace{-0.4ex}
    \captionof{figure}{A multi-robot collaborative depth perception experiment empowered by NeuroMesh framework with two aerial robots (green and blue) and one ground robot (red). The left column shows the input RGB images, the middle column shows the pointcloud represented in the consistent coordinate system. The right column shows the output depth image of each view.}
    \label{fig:teaser_strip}
    \vspace{-16pt}
}

\makeatother
\maketitle

\begin{abstract}
Deploying learned multi-robot models on heterogeneous robots remains challenging due to hardware heterogeneity, communication constraints, and the lack of a unified execution stack. This paper presents NeuroMesh, a multi-domain, cross-platform, and modular decentralized neural inference framework that standardizes observation encoding, message passing, aggregation, and task decoding in a unified pipeline. NeuroMesh combines a dual-aggregation paradigm for reduction- and broadcast-based information fusion with a parallelized architecture that decouples cycle time from end-to-end latency. Our high-performance C++ implementation leverages Zenoh for inter-robot communication and supports hybrid GPU/CPU inference. We validate NeuroMesh on a heterogeneous team of aerial and ground robots across collaborative perception, decentralized control, and task assignment, demonstrating robust operation across diverse task structures and payload sizes. We plan to release NeuroMesh as an open-source framework to the community.
\end{abstract}
\vspace{-10pt}
\section*{Supplementary material}
\textbf{Project / Code}:
\url{https://arplaboratory.github.io/NeuroMesh}

\section{Introduction}~\label{sec:introduction}
In recent years, robotics has witnessed a significant shift toward multi-robot systems, driven by the need for greater resilience, robustness, efficiency, and performance in complex environments. Learning-based multi-robot collaboration methods allow robots to share neural representations and computational resources to achieve objectives beyond the capability of individuals~\cite{pistilli2023graph, wu2024state}. Compared with single-robot systems, these approaches improve robustness to single-point failures, efficiency through task parallelization, and performance through collective intelligence.

The potential of learning-based multi-robot collaboration has been recognized across several domains. Applications include collaborative perception~\cite{zhou2022multi, li2023multirobot}, where robots exchange sensor information to improve scene understanding, cooperative control~\cite{tolstaya2020learning, chen2024learning, gosrich2021coverage, hu2020vgai}, where robots coordinate their actions to execute complex maneuvers, and planning~\cite{li2020graph, herrera2023learning, tzes2022graph, ji2021decentralized} and task assignment~\cite{ chebrolu2021graph, goarin2024graph, zhang2024scalable, wang2022heterogeneous}, where learned decentralized coordination improves resource allocation and mission execution. As these applications mature, there is growing interest in moving beyond simulations and datasets toward real-world deployment on physical robot teams.

However, transitioning these algorithms from theory to practice remains difficult. A practical system must integrate perception, control, and decision-making in a unified execution stack, support heterogeneous compute resources and inference engines, maintain robustness under decentralized communication, and expose interfaces that are flexible enough to accommodate task-dependent models. Existing works typically focus on a single task family, platform, or communication setting, leaving a gap between algorithmic demonstrations and reusable deployment infrastructure for heterogeneous multi-robot teams.

To address this gap, we present NeuroMesh, a unified decentralized neural inference framework for real-world multi-robot systems. NeuroMesh provides a flexible four-stage pipeline---observation encoding, message passing, aggregation, and task output decoding---with task-dependent encoder/decoder interfaces and a common execution stack rather than a new training method. We build NeuroMesh on three principles: generality (compatibility across multiple task domains and platforms), modularity (plug-and-play and scalable implementation), and decentralization (independent instantiation across robots).
While optimized for decentralized mesh networks, the framework remains compatible with centralized Wi-Fi topologies, ensuring broad accessibility for the robotics community.
The contributions of this paper are:
\begin{itemize}
\item \textbf{Multi-domain}: We propose a dual-aggregation paradigm that unifies dense perception and compact-message decentralized neural inference problems within a single formulation, enabling the instantiation and deployment of diverse task domains in one framework.
\item \textbf{Cross-Platform}: We demonstrate the framework's effectiveness across heterogeneous robotic platforms, with software modules that instantiate inference tasks on CPU or GPU to accommodate algorithmic needs and hardware constraints.
\item \textbf{Modularity}: The proposed framework adopts a plug-and-play microservice architecture in which task-dependent encoders, aggregators, and decoders can range from identity mappings to deep networks, while reusing the same parallel execution pipeline.
\item \textbf{Efficiency}: We provide a high-performance C++ implementation optimized for real-time operation, featuring an efficient network protocol implementation and an accelerated network inference engine implementation.
\end{itemize}
\section{Related Works}~\label{sec:relatedworks}
\vspace{-15pt}
\subsection{Learning-Based Multi-Robot Collaboration}
Learning-based multi-robot collaboration~\cite{wu2024state,pistilli2023graph,orr2023multi, li2023multi, liu2020when2com, liu2020who2com, hu2022where2comm} has recently gained significant attention. Among the works in this field, Graph Neural Networks (GNNs) have become a popular approach because they are decentralized and can naturally express the problem in a graph structure.

In multi-robot collaborative perception, \cite{zhou2022multi} utilizes GNNs for image-based multi-robot perception tasks including monocular depth estimation and semantic segmentation.
For LiDAR-based perception for autonomous driving scenarios,~\cite{li2023multirobot} introduces a scene-completion-based pretraining method.
Additionally, \cite{blumenkamp2023see} utilizes collaborative perception for multi-robot visual navigation.

Several works also adopt GNNs for decentralized multi-robot control.
\cite{tolstaya2020learning} presents a framework that enables robots to learn decentralized controllers using local information by imitating centralized controllers.
Similarly, \cite{chen2024learning} introduces a spatial-temporal GNN for decentralized flocking control.
Other works, such as \cite{gosrich2021coverage} and \cite{hu2020vgai}, focus on optimizing coverage control and vision-based coordination in robot swarms.

GNNs are also used to solve complex multi-robot path planning problems.
\cite{li2020graph} proposes a model that synthesizes local communication and decision policies for robot navigation in constrained environments.
Furthermore, \cite{tzes2022graph} and \cite{ji2021decentralized} demonstrate the effectiveness of GNNs in active information acquisition and navigation.
For task assignment,~\cite{chebrolu2021graph} proposes a learning-based distributed approach to enhance traditional auction and consensus-based algorithms.
\cite{goarin2024graph} presents a decentralized GNN-based goal assignment strategy for a robot team in scenarios with restricted communication.
Furthermore, \cite{zhang2024scalable} employs graph-based deep reinforcement learning to generate the probability of each task being assigned to each robot.
Finally, \cite{wang2022heterogeneous} explores multi-robot task assignments using a heterogeneous graph attention network.

\vspace{-5pt}
\begin{figure*}[t]
\centering
\includegraphics[width=0.9\linewidth, trim=0 8 0 0, clip]{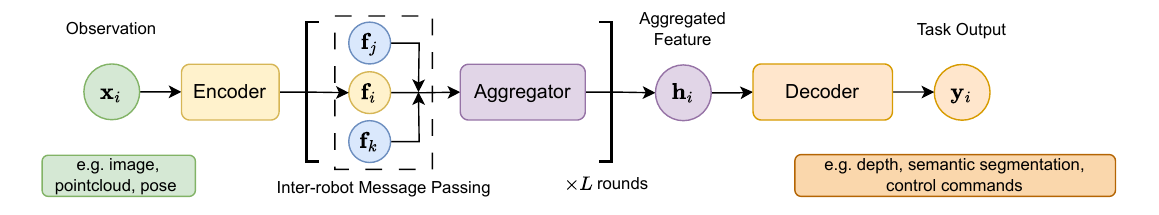}
\caption{NeuroMesh four-stage process modeling of decentralized multi-robot neural inference for a generic robot $i$. The encoded feature $\mathbf{f}_i$ and received encoded features ($\mathbf{f}_j, \mathbf{f}_k$, etc.) from connected robots are passed into the message passing and aggregation stage. The message passing and aggregation stage can be executed for $L$ rounds. In the decoder stage, the aggregated feature $\mathbf{h}_i$ is transformed into a tensor and produces the task output $\mathbf{y}_i$.}
\label{fig:methodology}
\vspace{-10pt}
\end{figure*}

\subsection{Frameworks for Multi-Robot Collaboration}
While most learning-based multi-robot collaboration works demonstrate their capabilities in simulation and on datasets, there is a gap in real-world deployment.
Several frameworks try to bridge this gap. 
ModGNN~\cite{korvesely2021modgnn} focuses on architectural modularity for multi‑agent control but does not provide a deployment stack with inter‑robot communication, hardware‑accelerated inference, or ROS~2 microservices for real‑time operation on heterogeneous robots.
The ROS~2 framework of~\cite{blumenkamp2022framework} demonstrates real‑robot decentralized control but is Python‑based, tailored to a single task family (control) and platform‑specific payloads; its computation/communication stack is not optimized for perception‑heavy or mixed‑domain workloads.

NeuroMesh addresses the practical deployment gap for decentralized multi-robot neural inference with a multi-domain, cross-platform, modular, and high-performance framework.
\vspace{-5pt}
\section{Methodology}~\label{sec:methodology}
\vspace{-15pt}

The decentralized multi-robot neural inference problem, as shown in Fig.~\ref{fig:methodology}, can be formulated as a four-stage process: \textbf{observation encoding}, \textbf{message passing}, \textbf{aggregation}, and \textbf{task output decoding}.
In the \textbf{observation encoding} stage, each robot $i$ encodes its local observations $\mathbf{x}_i$ (e.g., image, point cloud, poses, etc.) into a feature tensor $\mathbf{f}_i$.
During the \textbf{message passing} stage, robots exchange information with their neighbors within communication range. We denote $\mathcal{N}(i)$ as the set of robots connected to robot $i$. Subsequently, in the \textbf{aggregation stage}, each agent aggregates the received information from neighboring agents to update its feature tensor $\mathbf{f}_i$, obtaining $\mathbf{h}_i$.
In the \textbf{task output decoding} stage, each robot first decodes its feature tensor $\mathbf{h}_i$ to obtain the task output $\mathbf{y}_i$ (e.g., depth, semantic, control commands, etc.).

The problem formulation for $L$ communication rounds is
\begin{equation}
\label{eq:problem_formulation}
\begin{aligned}
\mathbf{f}_i &= \text{Encode}(\mathbf{x}_i), \\
\mathbf{h}_i^{(0)} &= \mathbf{f}_i, \\
\mathbf{h}_i^{(l+1)} &= \text{Aggregate}(\mathbf{h}_i^{(l)}, \underbrace{\{\mathbf{h}_j^{(l)} : j \in \mathcal{N}(i)\}}_\text{Message Passing}), \\
l &= 0, \cdots, L-1 \\
\mathbf{y}_i &= \text{Decode}(\mathbf{h}_i^{(L)}).
\end{aligned}
\end{equation}
In Section~\ref{sec:casestudy}, we list several examples to illustrate how the proposed formulation addresses multiple task types. The encoder, message passing, aggregation, and decoder modules are task-dependent and can be selected by the user to match the specific performance and hardware requirements. In particular, the encoder/decoder pair may be a deep vision stack for perception or a lightweight learned mapping for compact state/cost vectors in control and task assignment.

\begin{figure}[t]
    \centering
    \begin{subfigure}[b]{\linewidth}
        \centering
        \includegraphics[width=0.7\linewidth]{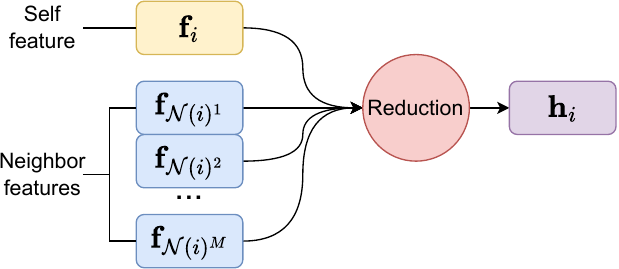}
        \caption{Reduction aggregation paradigm}
        \label{fig:aggregation_reduction}
    \end{subfigure}
    \begin{subfigure}[b]{\linewidth}
    \centering
    \includegraphics[width=0.7\linewidth]{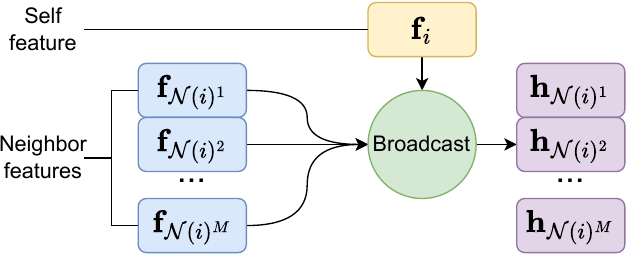}
    \caption{Broadcast aggregation paradigm}
    \label{fig:aggregation_broadcast}
    \end{subfigure}    
    \caption{Two aggregation paradigms. In Fig. \ref{fig:aggregation_reduction}, the neighbor features in blue aggregate with self features $\mathbf{f}_i$ in yellow and reduce to aggregated features $\mathbf{h}_i$ in purple. In Fig. \ref{fig:aggregation_broadcast}, the self feature $\mathbf{f}_i$ in yellow is broadcasted to neighbor features in blue, to obtain aggregated feature $\mathbf{h}_i^m$ for each neighbor feature and assembled to $\mathbf{h}_i$.}
    \vspace{-10pt}
    \label{fig:aggregation}
\end{figure}
\begin{figure}[!ht]
    \centering
\includegraphics[width=0.9\linewidth, trim=0 215 0 0, clip]{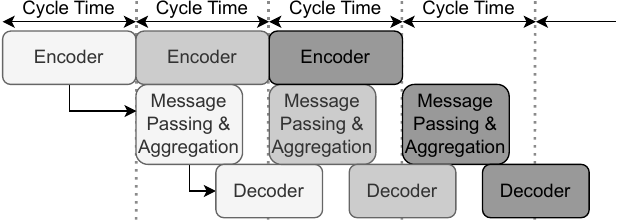}
    \caption{The parallel three-stage pipeline. The stages run concurrently, allowing the Encoder to process data for cycle $N$ while the Aggregator and Decoder handle data from cycles $N-1$ and $N-2$, respectively. The system's output cycle time is determined by the duration of the slowest stage, not the sum of all stages, significantly improving throughput. 
    }
    \label{fig:NeuroMesh_cycle}
    \vspace{-10pt}
\end{figure}
A key contribution of NeuroMesh is the unification of multiple collaboration domains through a dual-aggregation formulation. Existing formulations typically rely on a ``Reduction'' paradigm (e.g., $\sum$ or max-pooling). While sufficient for consensus-like control tasks, this approach destroys the pairwise relational information required for dense perception tasks.
We introduce a dual-aggregation paradigm, as shown in Fig.~\ref{fig:aggregation}, enabling the system to execute algorithms ranging from simple state synchronization to complex multi-view geometric reasoning without architectural redesign. The reduction paradigm reduces the self-feature $\mathbf{f}_i$ and neighbor features $\{\mathbf{f}_j : j \in \mathcal{N}(i)\}$ to one aggregated feature $\mathbf{h}_i$. For example, we can compute the average of the self-feature and the neighbor features to obtain the aggregated feature.
The Broadcast paradigm computes pairwise interactions. For $M$ neighbors, the self-feature $\mathbf{f}_i$ is replicated and concatenated with each neighbor feature $\mathbf{f}_j$. A network then processes these pairs in parallel to produce an aggregated tensor $\mathbf{h}_i$ with $M$ channels, preserving relational information.
In the event of intermittent communication or message drops from individual robots, the framework can either pause operations until the required messages are received or dynamically adapt by transitioning to a partial-neighborhood or single-robot mode through adjustment of the aggregation step. This choice depends on whether the task can still yield a valid output with incomplete neighbor information.

Real-world deployment faces the challenge of computational latency accumulating across complex perception and control stacks. To decouple the cycle time from the total processing time, we design a parallel three-component pipeline, illustrated in Fig.~\ref{fig:NeuroMesh_cycle}.
The three components are the encoder module, the message passing and aggregation module, and the decoder module. The parallel pipeline reduces system latency and improves throughput compared to sequential computation.

To save computational resources, we use the encoder cycle time as the pipeline rate, since the encoder callback drives the entire pipeline.
Let $T_e, T_m, \text{and } T_d$ denote the computation times for the encoder, message passing, and decoder callbacks, respectively. The pipeline output cycle time is $T_{\text{pipeline}} = \max(T_e, T_m, T_d)$, and the delay of the entire pipeline is $T_{\text{delay}} = T_e + T_m + T_d$.
This approach reduces the output cycle time of the system and improves the overall system throughput compared to sequential computation.

\section{Architecture}

\begin{figure}[t]
\centering
\includegraphics[width=0.8\linewidth,trim=0 9 0 0, clip]{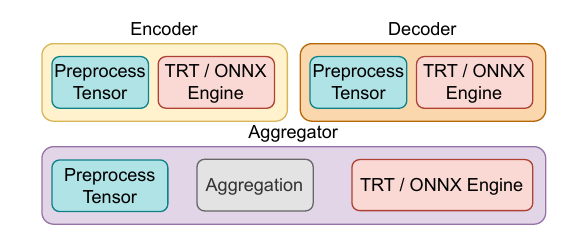}
\caption{The instantiation of NeuroMesh modules described in Fig.~\ref{fig:methodology}. The encoder employs TRT/ONNX engine execution or ONNX. The aggregator stage combines features after the message passing and can be optionally passed into the engine. The decoder also employs the same engine.}
\label{fig:software_system_architecture}
\vspace{-10pt}
\end{figure}

In this section, we present the approach's implementation details. We describe how the encoder, aggregator, and decoder in Fig.~\ref{fig:methodology} are instantiated into our proposed framework. The system architecture is illustrated in Fig.~\ref{fig:software_system_architecture}.

\subsection{Software Implementation}
Our framework adopts TensorRT (TRT)~\cite{TensorRT} as the neural network inference engine for the NVIDIA platform. We also support ONNX Runtime for non-NVIDIA platforms as an option. During the initialization stage, the framework loads the TensorRT or ONNX weights of the subnetworks and initializes the encoder module, aggregator module, and decoder module of the pipeline. Subsequently, the framework awaits observation data to trigger the pipeline. The framework is agnostic to the specific neural network design. The user can flexibly insert task-dependent neural networks in a plug-and-play manner, as demonstrated in Section~\ref{sec:casestudy}.

The encoder module preprocesses the input observation into tensor format and applies the encoder network to the observation data $\mathbf{x}_i$ using the TensorRT or ONNX engine to obtain the feature tensor $\mathbf{f}_i$ of robot $i$.
The message-passing module exchanges encoded feature tensors through the inter-robot communication layer across the multi-robot team.
The aggregator module preprocesses the message data into tensor format and applies a problem-specific aggregation function from the aforementioned Reduction/Broadcast aggregation paradigm to the feature tensor $\mathbf{f}_i$ of the robot and the tensors received from its neighbors. The output of the aggregation module is the updated feature tensor $\mathbf{h}_i$, computed using the TensorRT or ONNX engine.
The decoder module preprocesses the aggregated feature data into tensor format and applies the task output decoding network using the TensorRT or ONNX engine to the updated feature tensor $\mathbf{h}_i$ to obtain the task output tensor $\mathbf{y}_i$ of the robot.

We implement NeuroMesh in ROS\,2~\cite{macenski2022robot}. Each stage is implemented as a ROS\,2 wrapper node built on an optimized C++ backend to minimize serialization overhead and maximize inference throughput. ROS\,2 employs a microservice-based architecture with nodes as computational units that communicate through topics using a publisher-subscriber model. Our framework adopts the microservice architecture supported by ROS\,2, where each component is implemented as a ROS\,2 callback and uses ROS\,2 topics for inter-robot and intra-robot communications.

This architecture separates task-specific model design from system-level execution. Local observations enter a task-defined encoder callback, encoded tensors are published locally and over the network, the aggregator consumes the current self-feature together with buffered neighbor features, and the decoder emits the task output. The same runtime therefore supports different tasks under a common interface.

\subsection{Communication Protocol}\label{sec:communication_protocol}
The proposed framework design is agnostic to centralized or decentralized network topology. In the context of multi-robot system deployment, the communication cost can easily surpass the bandwidth limit of centralized network infrastructure as the number of robots increases. Therefore, a decentralized network topology is preferred over a centralized network topology. We use ROS\,2 for intra-robot communication within this framework. We found that its native middleware implementations, such as Fast RTPS and Cyclone DDS, cannot satisfy the communication requirements of specific applications, as demonstrated in the results presented in Section~\ref{sec:casestudy}.
Therefore, we configure native ROS\,2 DDS middleware implementations and incorporate Zenoh~\cite{corsaro2023zenoh}, a communication protocol that combines publish/subscribe primitives, geo-distributed storage, and querying.
In Zenoh peer mode, each robot exchanges traffic directly with other peers or uses link-state routing. In Zenoh router mode, one or more Zenoh routers forward traffic between connected nodes.
Each inter-robot message carries a timestamp and sequence index. Every robot maintains a keep-latest buffer for each neighbor, discards out-of-order packets, and removes messages older than a task-defined threshold $\Delta t$. Aggregation can operate either in blocking mode, waiting for all required messages, or in best-effort mode, using the most recent subset of neighbors and falling back to partial-neighborhood or single-robot inference when the task semantics permit it.
\vspace{-5pt}
\section{Case Studies}~\label{sec:casestudy}
We demonstrate the performance of the proposed system in three distinct multi-robot collaboration tasks---perception, control, and task assignment---highlighting the framework's compatibility across various tasks and platforms.

These case studies validate NeuroMesh's core claims: multi-domain applicability, cross-platform deployment, modularity, and efficiency by testing NeuroMesh across high-bandwidth perception, high-frequency control, and discrete task assignment. We use the same pipeline on a heterogeneous aerial-ground team, swapping only task-specific modules; perception/control use 3 robots and task assignment uses 5. Although the encoder interface is modality-agnostic, the perception case studies in this paper use RGB inputs.

We employ a team of ground and aerial robots equipped with Doodlelabs mini Mesh Rider Radios\footnote{\url{https://doodlelabs.com/product/mini-mesh-rider-radio/}} for a decentralized mesh-network setting. Each robot can communicate with its neighbors within the same mesh network.
Aerial robots are equipped with an NVIDIA Jetson Orin NX as the computation platform, whereas the three ground robots are equipped with Mini-ITX computers with 4th-generation Intel i7 CPUs and NVIDIA 1050 Ti GPUs.

\subsection{Multi-Robot Collaborative Perception}
\begin{figure}[ht!]
    \centering
    \begin{subfigure}[b]{\linewidth}
        \centering
        \includegraphics[width=0.32\linewidth]{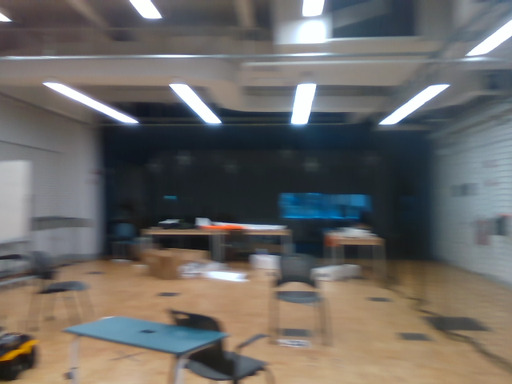}
        \includegraphics[width=0.32\linewidth]{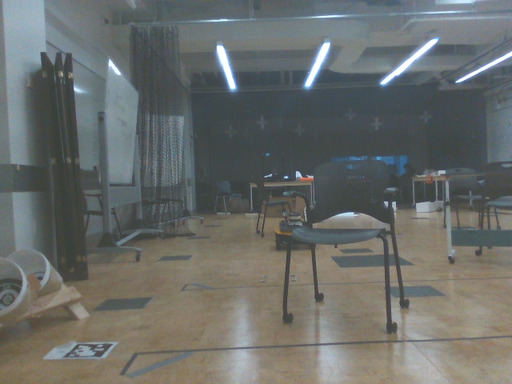}
        \includegraphics[width=0.32\linewidth]{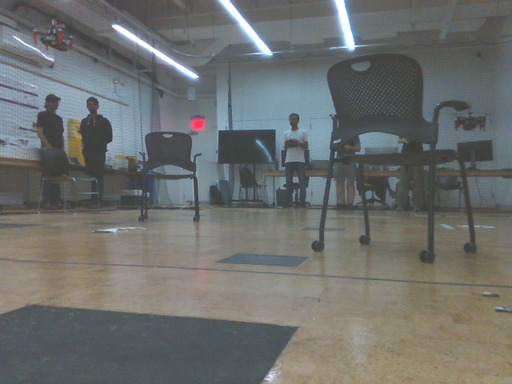} \\
        \vspace{0.6ex}
        \includegraphics[width=0.32\linewidth]{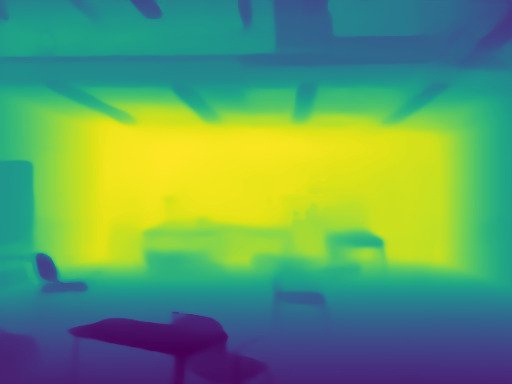}
        \includegraphics[width=0.32\linewidth]{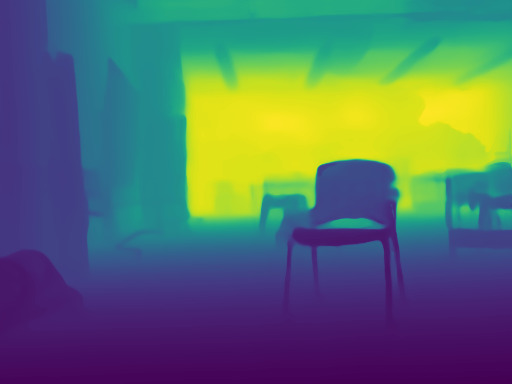}
        \includegraphics[width=0.32\linewidth]{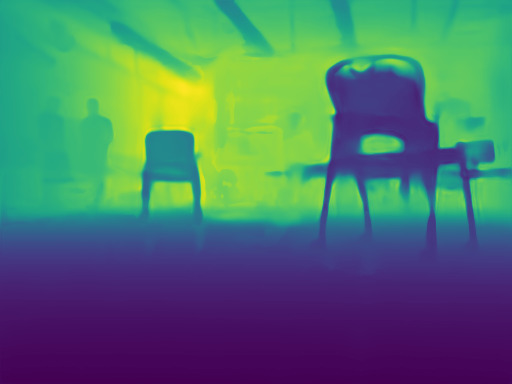} \\
        \vspace{2ex}
        \includegraphics[width=0.32\linewidth]{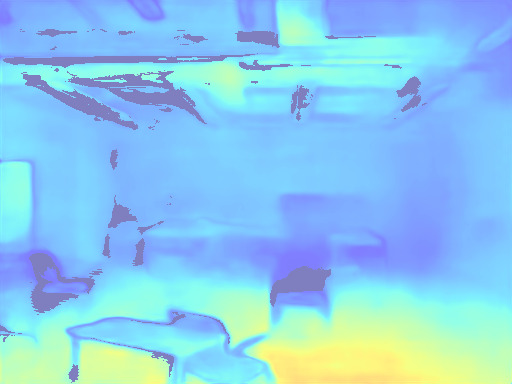}
        \includegraphics[width=0.32\linewidth]{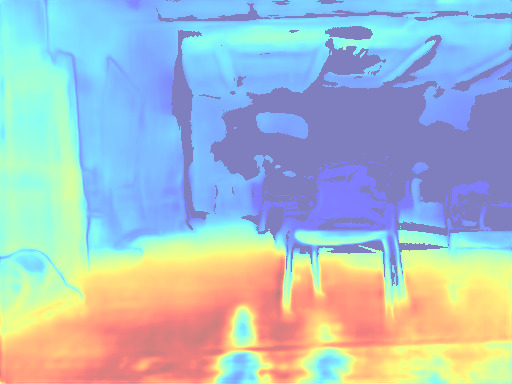}
        \includegraphics[width=0.32\linewidth]{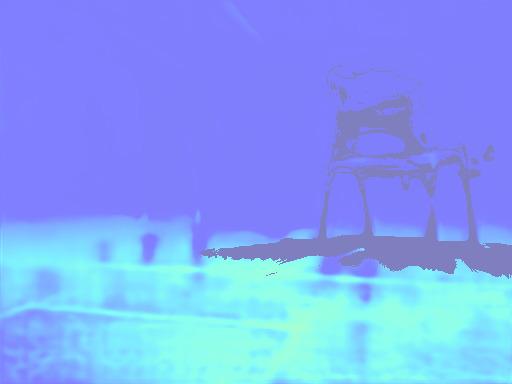}
        \makebox[0.32\linewidth]{Aerial robot} \makebox[0.32\linewidth]{Aerial robot} \makebox[0.32\linewidth]{Ground robot}
        \caption{\label{fig:perception_a}} 
    \end{subfigure}

    \begin{subfigure}[b]{\linewidth}
        \centering
        \includegraphics[width=0.32\linewidth]{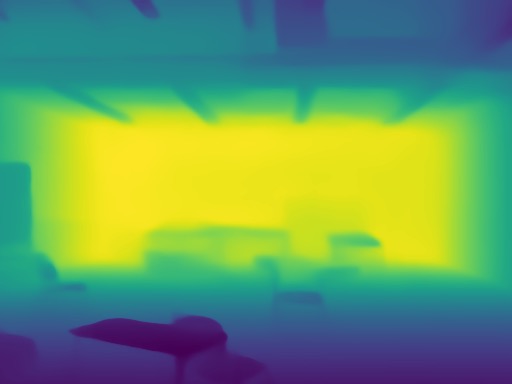}
        \includegraphics[width=0.32\linewidth]{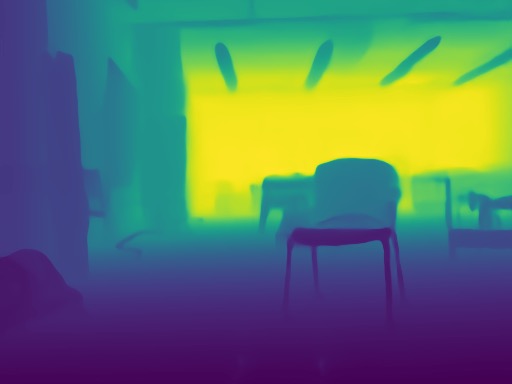}
        \includegraphics[width=0.32\linewidth]{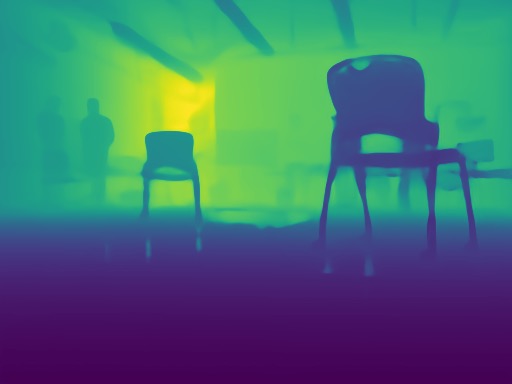} \\
        \vspace{2ex}
        \includegraphics[width=0.32\linewidth]{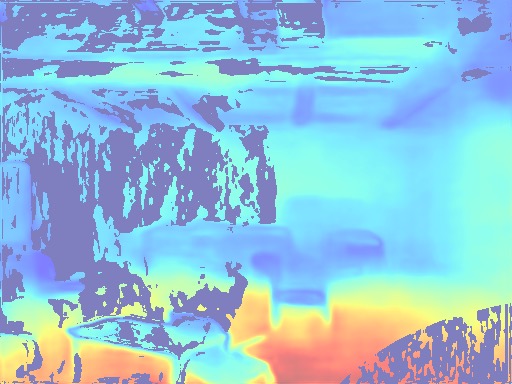}
        \includegraphics[width=0.32\linewidth]{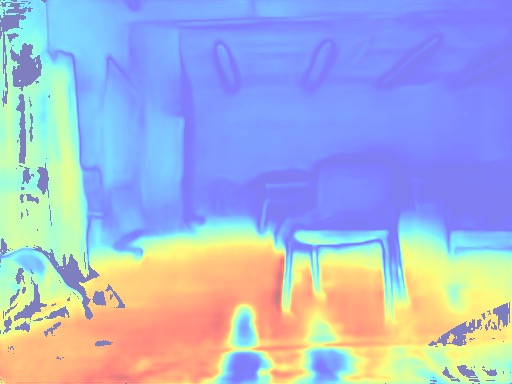}
        \includegraphics[width=0.32\linewidth]{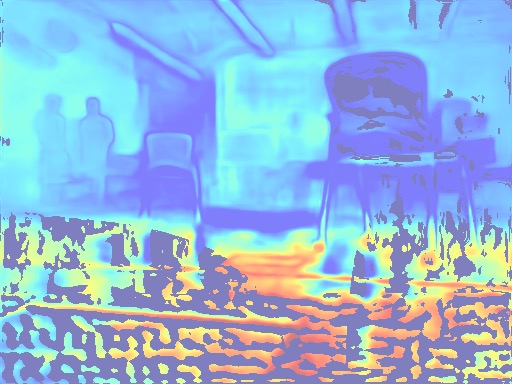}
        \makebox[0.32\linewidth]{Aerial robot} \makebox[0.32\linewidth]{Aerial robot} \makebox[0.32\linewidth]{Ground robot}
        \caption{\label{fig:perception_as}} 
    \end{subfigure}
    
    \vspace{0.6ex}
    \begin{subfigure}[b]{\linewidth}
        \centering
        \includegraphics[width=0.32\linewidth]{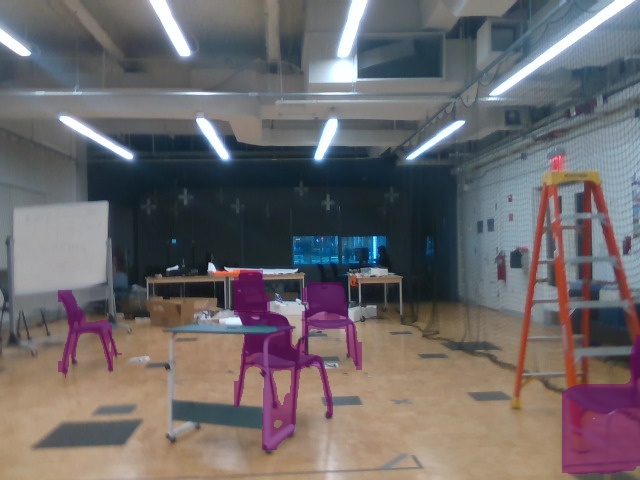}
        \includegraphics[width=0.32\linewidth]{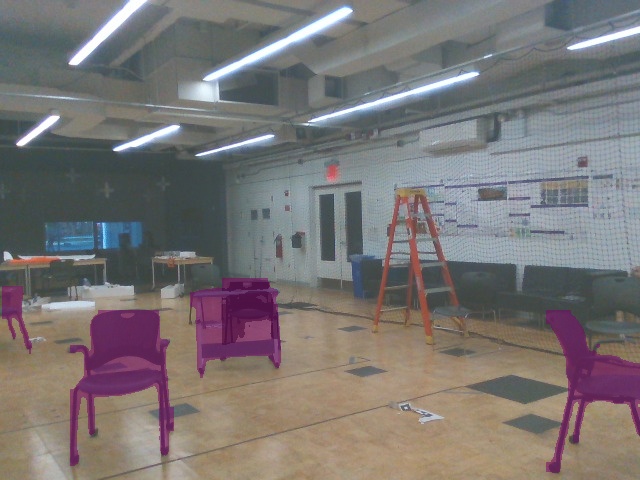}
        \includegraphics[width=0.32\linewidth]{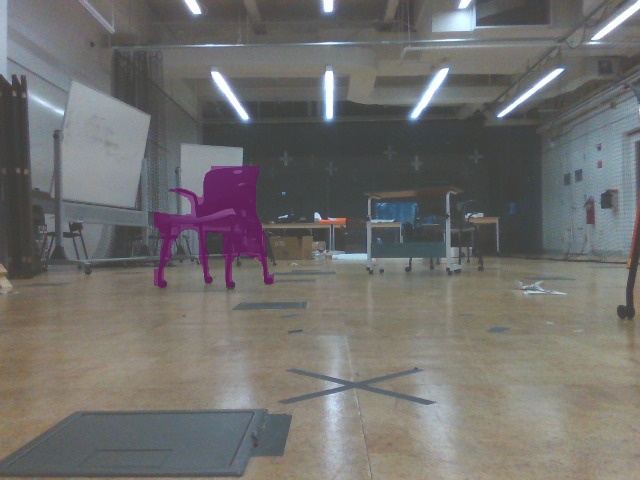}
        \makebox[0.32\linewidth]{Aerial robot} \makebox[0.32\linewidth]{Aerial robot} \makebox[0.32\linewidth]{Ground robot}
        \caption{\label{fig:perception_b}} 
    \end{subfigure}
    
    \caption{Multi-Robot Collaborative Perception. Fig.~\ref{fig:perception_a} shows collaborative depth estimation by DUSt3R~\cite{wang2024dust3r}. The rows from top to bottom are input RGB image, depth prediction, and model-predicted uncertainty. Fig.~\ref{fig:perception_as} shows the corresponding single-robot monocular depth estimation and uncertainty baselines using the same images. The uncertainty maps are obtained from the DUSt3R single-forward confidence output, clamped to a fixed range, and normalized with the same mapping for both settings; lower uncertainty appears in \textcolor{blue}{blue} and higher uncertainty in \textcolor{red}{red}. Fig.~\ref{fig:perception_b} shows the collaborative semantic segmentation of chair category by \cite{zhou2022multi}. \label{fig:perception_qualitative}}
    \vspace{-10pt}
\end{figure}

We showcase two examples of multi-robot collaborative perception with the NeuroMesh framework on a heterogeneous robot team of two aerial robots and one ground robot. Qualitative perception results are presented in Fig.~\ref{fig:perception_qualitative}.

First, we design a decentralized version of DUSt3R~\cite{wang2024dust3r} for collaborative depth estimation. A Vision Transformer encodes RGB images, encoded features are exchanged among robots, broadcast aggregation pairs self and neighbor features through the transformer decoder, and the head regresses depth from the resulting point maps. We use the weights from the original manuscript \cite{wang2024dust3r} and convert the model to a decentralized version using our setup. The uncertainty visualization in Fig.~\ref{fig:perception_qualitative} is obtained from the DUSt3R single-forward confidence output, represented as a learned per-pixel log-variance; we clamp and linearly normalize it with the same mapping for the single-robot and collaborative results. The single-robot baseline in Fig.~\ref{fig:perception_as} shows higher uncertainty and less accurate depth than the collaborative result in Fig.~\ref{fig:perception_a}. We evaluate depth accuracy using Absolute Relative Error and Inlier Ratio with a threshold of $1.03$ following \cite{schroppel2022benchmark} on our collected indoor / outdoor dataset.
Both centralized and decentralized setups obtain the same absolute relative error of $5.29$ and inlier ratio of $61.32$.
We conclude that centralized inference and decentralized inference with NeuroMesh produce equivalent results.

Second, we deploy a decentralized GNN-based collaborative semantic segmentation method~\cite{zhou2022multi}. CNN encoders transform RGB images into features, message passing exchanges them among neighbors, reduction aggregation fuses them, and CNN decoders regress semantic segmentation images.
In our qualitative results shown in Fig.~\ref{fig:perception_b}, the collaboration helps each robot recognize chairs from different perspectives.

\subsection{Multi-Robot Collaborative Control}
\begin{figure}[t]\centering\includegraphics[width=0.8\linewidth, trim = 0 15 0 12, clip]{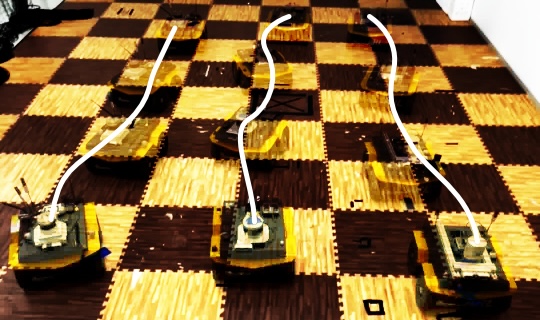}
    \caption{Multi-Robot collaborative control. Three ground robots perform point-to-point navigation without collision. }
    \label{fig:control_quali}
    \vspace{-10pt}
\end{figure}
We use the NeuroMesh framework to deploy multi-robot control policies on three ground robots with non-holonomic dynamic constraints.
Specifically, we train an RL-based policy for multi-robot point-to-point navigation.
The three ground robots need to reach their corresponding target positions without colliding with each other.
To achieve this, we modify the GNN-based approach from \cite{blumenkamp2022framework}, originally designed for holonomic platforms, to suit our needs by adjusting it to a unicycle motion model.

Inspired by \cite{blumenkamp2022framework}, we train a GNN-based policy using reinforcement learning. The reward design encourages the robots to reach their designated goals in the shortest time and penalizes collisions.
Each robot obtains its pose and heading estimate from onboard localization (e.g., SLAM/VIO), and NeuroMesh exchanges only compact state features rather than raw localization sensor streams. The framework does not require motion-capture infrastructure and can also accommodate other localization sources such as UWB or OptiTrack when available. A four-layer MLP encodes observations $\mathbf{o}_i$ into features $\mathbf{f}_i$, message passing shares them within the team, aggregation processes neighbor differences, and a decoder extracts the policy.
The agent's observation is given by
\begin{equation}
    \mathbf{o}_i = (\mathbf{g}_i - \mathbf{x}_i, \mathbf{x}_i, \mathbf{u}(\theta_i), \mathbf{x}_i + v_f \mathbf{u}(\theta_i)) \in \mathbb{R}^8,
    \vspace{-5pt}
\end{equation}
where $\mathbf{x}_i$ is the agent's 2D position, $\mathbf{g}_i$ is the goal position, $\mathbf{u}(\theta_i) = (\cos \theta_i, \sin \theta_i)$ is its planar heading, and $v_f$ is its forward velocity. Each agent applies a four-layer MLP to encode a feature map $\mathbf{f}_i=\text{Encoder}(\mathbf{o}_i)$ to be shared.
In the message passing stage, for each other agent $j \neq i$, a 3-layer MLP network $\bm g$ is applied to the difference $\mathbf{f}_j - \mathbf{f}_i$, and all messages are accumulated via
\begin{equation}
    \mathbf{h}_i = \sum_j \bm g(\mathbf{f}_{j} - \mathbf{f}_{i} ).
    \vspace{-5pt}
\end{equation}
In the decoding stage, a 4-layer network provides the raw output $\mathbf{y}_i = \text{Decoder}(\mathbf{h}_i) \in \mathbb{R}^4$.
We then apply a shifted ReLU entrywise $\tilde{\mathbf{y}}_i = \log(e^{\mathbf{y}_i} + 1)+1$.
The final output $\tilde{\mathbf{y}}_i$ parameterizes the Beta distributions from which forward and angular velocities are then sampled.

While the policies achieved an $80\%$ success rate in 2D motion simulations, matching the training environment of \cite{blumenkamp2022framework}, real-world deployment revealed significant sim-to-real transfer challenges, particularly in angular velocity control.
Our framework enabled real-time distributed execution of the GNN-based RL policies in the real world.
Through real-time corrections of errant angular velocities, we achieved a $60\%$ success rate in the three-robot navigation task. The task is successfully accomplished as shown in Fig.~\ref{fig:control_quali}.
\vspace{-5pt}
\subsection{Multi-Robot Collaborative Task Assignment}
We showcase a goal assignment task where $5$ robots must each select a goal within the environment while avoiding conflicting assignments. The objective is to find the optimal robot-to-goal assignment that minimizes a global cost metric, while satisfying a one-to-one matching constraint. This problem is commonly known as the linear sum assignment problem solved centrally by the Hungarian algorithm \cite{kuhn1955hungarian}. We minimize $\sum_{i,j} c_{ij} s_{ij}$ subject to one-to-one matching constraints, where $c_{ij}$ is the cost of assigning robot $i$ to goal $j$ and $s_{ij} \in \{0,1\}$ is the assignment label. In contrast to the perception setting, there is no raw sensor input here: the local observation is a compact cost vector computed from each robot's state estimate and the candidate goal set.

Inspired by \cite{goarin2024graph}, we train a GNN with supervised learning using the Hungarian algorithm as the expert. We frame the task as a classification problem where the GNN predicts a vector of assignment labels $\mathbf{s}=\begin{bmatrix}
    s_{i0}, \cdots, s_{i5}
\end{bmatrix}^\top$ for each robot $i$, indicating the goal ID to which it is assigned. The input feature of each robot is a vector of the costs of all possible assignments
$
    \mathbf{c}_{i} = \begin{bmatrix} c_{i0}, \cdots, c_{i5} \end{bmatrix}^\top.
$

In NeuroMesh, a 2-layer MLP encodes the cost vector, message passing shares feature embeddings, a graph-attention aggregator with $3$ heads and $2$ layers fuses them, and a lightweight decoder outputs assignments. This case study illustrates that the encoder/decoder can remain lightweight while the framework still provides value through decentralized transport, buffering, and graph-based aggregation. We physically validate this task outdoors with a 5-robot team, which is the largest hardware experiment in this paper.

We demonstrate the performance of this task by considering $N_{\text{tests}} = 20$ experiment runs.
We introduce two evaluation metrics:
the Success Rate ($SR$) and the Total Cost Performance ($TCP$). The $SR$ represents the percentage of goals covered once the robots have reached their assigned destinations
\begin{equation}
    SR = \frac{N_{goals}}{5 \times N_{tests}} \times 100,
\end{equation}
with $N_{goals}$ the total number of goals reached, whereas the $TCP$ measures the average percentage increase in cost relative to the optimal cost
\begin{equation}
    \text{TCP} = \frac{1}{N_{tests}} \sum_n \frac{C_{\text{GNN},n} - C_{\text{opt},n}}{C_{\text{opt},n}} \times 100,
\end{equation}
with $C_{\text{GNN},n}$ and $ C_{\text{opt},n}$ as the total costs obtained by the GNN and the expert respectively for $n \in [0, N_{\text{tests}}]$. The TCP metric compares the cost performance of the GNN and the expert and is only calculated when all goals are covered.

\begin{table}[t]
\centering
\caption{Goal assignment performance.}
\label{tab:ablation_goal_assignment}
\begin{tabular}{@{}lcccc@{}}
\toprule
\textbf{Message Size (Bytes)} & 40 & 64 & 128 & 256 \\
\midrule
\textbf{SR (\%)} & 87.12 & 90.22 & 93.92 & 94.01 \\
\midrule
\textbf{TCP (\%)} & 0.05 & 0.03 & 0.03 & 0.02 \\
\bottomrule
\end{tabular}
\vspace{-10pt}
\end{table}

We evaluate the scalability of the framework in terms of communication message size, ranging from $40$ B to $256$ B. In Table~\ref{tab:ablation_goal_assignment}, the total cost for successful runs is nearly optimal, at most $0.05\%$ higher than the expert's optimal cost. However, message size significantly affects the success rate. Larger message sizes improve performance, highlighting a trade-off between message size and success rate. A message size of $128$ B is sufficient to achieve near-optimal results, with a success rate of $93.92\%$, while increasing the size to $256$ B provides no further performance gain. The total cost for successful runs is nearly optimal, only $0.03\%$ higher than the expert's optimal cost.

\vspace{-5pt}
\section{Performance Analysis}~\label{sec:result}
\vspace{-5pt}

\subsection{Communication Ablation Study}
\begin{figure}[t]
    \centering
    \begin{subfigure}[b]{\linewidth}
        \centering
        \includegraphics[width=\linewidth, trim=8 20 5 5, clip]{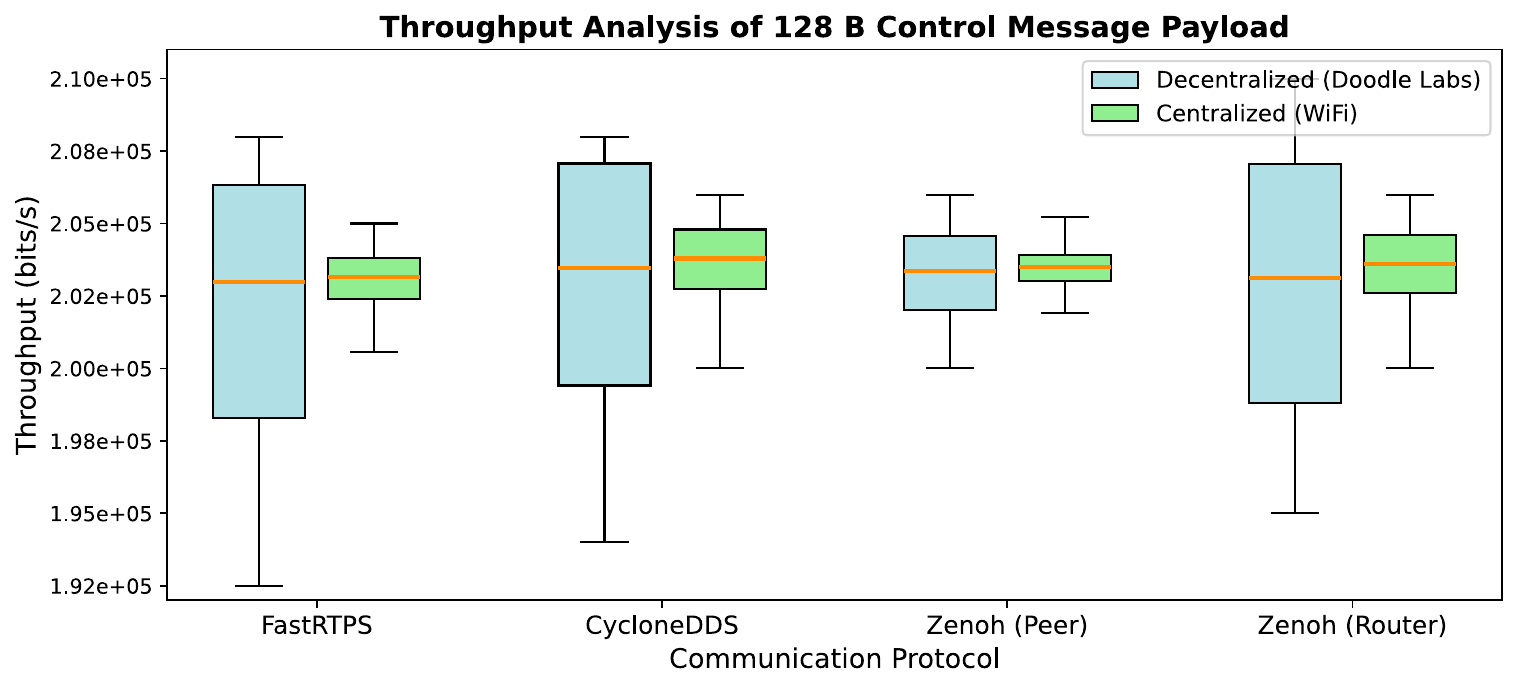}
    \end{subfigure}
    \begin{subfigure}[b]{\linewidth}
        \centering
        \includegraphics[width=\linewidth,trim=8 25 5 5, clip]{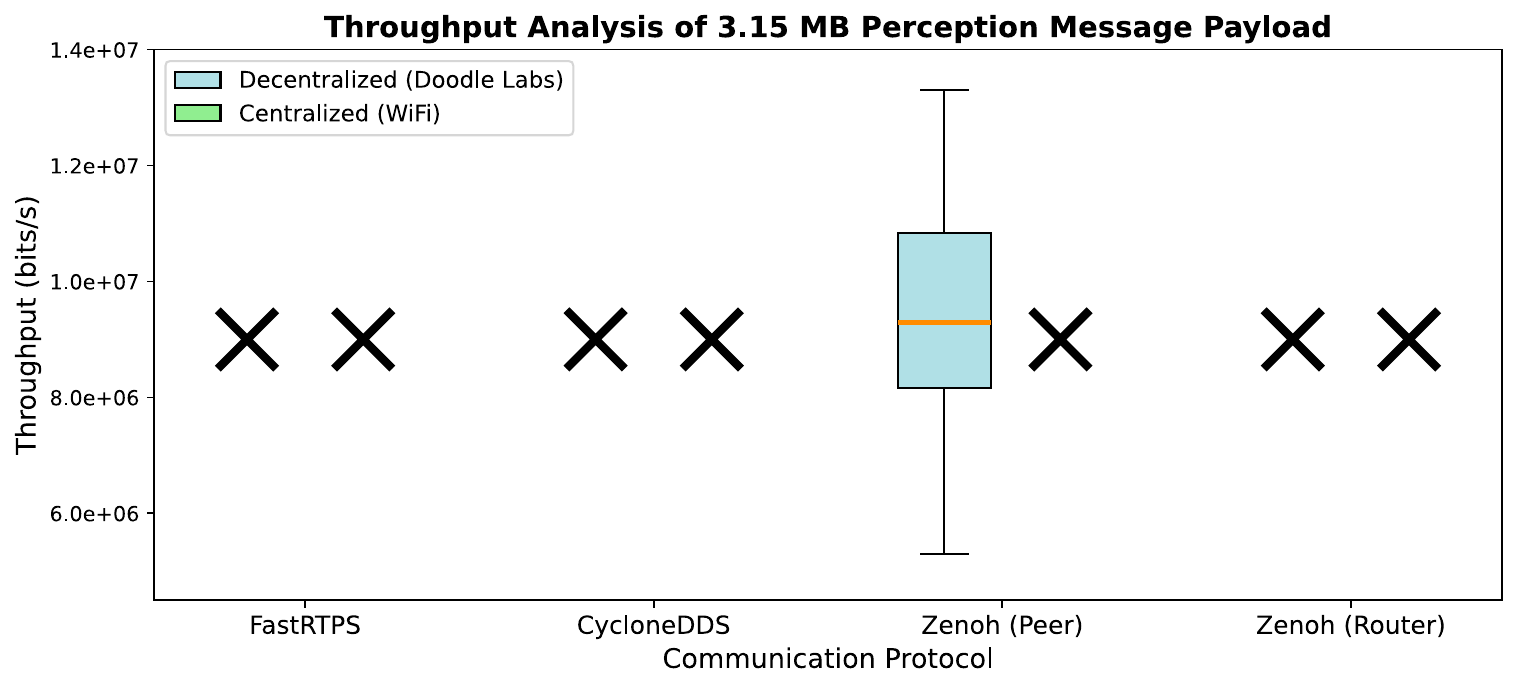}
    \end{subfigure}
    \caption{Throughput analysis of small control message payload (top) and large perception message payload (bottom). 
    }
    \label{fig:communication}
\end{figure}

NeuroMesh supports both decentralized mesh through Doodlelabs mesh radios and centralized Wi-Fi topologies with four transport protocols (Fast RTPS, CycloneDDS, Zenoh peer, and Zenoh router mode) presented in Section~\ref{sec:communication_protocol}.
We evaluate communication performance on both setups in representative indoor lab/office conditions with robots in motion. Live pipelines generate two payloads: a $128$ B control payload and a $3.15$ MB perception payload. Fig.~\ref{fig:communication} compares throughput across four protocols with both decentralized (Doodlelabs mesh radios) and centralized network setups (Wi-Fi), and Table~\ref{tab:communication_quality} reports end-to-end communication quality metrics for the decentralized network setup.

When operating Doodlelabs mesh radios, Fast RTPS, CycloneDDS, and Zenoh router mode carry control messages but exhibit fluctuating throughput (ranging roughly between $50$ and $200~\si{Hz}$) for the $128$ B control payload. Zenoh peer mode achieves a stable $200~\si{Hz}$ throughput. It also exhibits the lowest latency ($4.8$~ms), jitter ($0.6$~ms), and packet loss ($0.3\%$) among transports (Table~\ref{tab:communication_quality}).
For the $3.15$ MB perception payload, Fast RTPS, CycloneDDS, and Zenoh router mode fail on the mesh. Only Zenoh peer mode succeeds, achieving approximately $0.5~\si{Hz}$. Although latency is substantially higher ($2712.2$ ms, jitter $540.5$ ms) due to the large serialized tensor, Zenoh peer mode is the only protocol that makes decentralized collaborative perception feasible on the mesh network. Overall, the Zenoh-ROS2 bridge in peer mode combined with a decentralized Doodlelabs mesh network provided the most robust throughput for the decentralized setup.

We also evaluate a centralized 802.11 Wi-Fi deployment (Ubiquiti Networks' airMax Nanostation M5\footnote{\url{https://dl.ubnt.com/datasheets/nanostationm/nsm_ds_web.pdf}}). On the $128$ B control task, all four transport protocols achieve similar throughput ranges, while Zenoh peer mode achieves the highest throughput stability.
However, we could not sustain the $3.15$ MB perception payload with centralized Wi-Fi, as routing large tensors through the central node saturated the available bandwidth.
Overall, Zenoh router mode over centralized Wi-Fi is a viable alternative for control-class workloads in deployments without mesh radio infrastructure.
\begin{table}[t]
\centering
\caption{End-to-end communication quality metrics for the real-robot communication ablation runs with $3$ robots, $1$--$7$~m separation, and moderate indoor RF interference.}
\label{tab:communication_quality}
\setlength{\tabcolsep}{1pt}
\resizebox{\columnwidth}{!}{%
\begin{tabular}{@{}lllccc@{}}
\toprule
Hardware & Scenario & Transport & Latency (ms) & Jitter (ms) & Packet Loss (\%) \\
\midrule
Mesh & Control & Fast RTPS & 8.6 & 2.9 & 2.1 \\
Mesh & Control & CycloneDDS & 6.7 & 1.7 & 1.2 \\
Mesh & Control & Zenoh peer & 4.8 & 0.6 & 0.3 \\
\midrule
Mesh & Perception & Zenoh peer & 2712.2 & 540.5 & 10.9 \\
\bottomrule
\end{tabular}
}
\vspace{-10pt}
\end{table}

To assess scaling beyond the physical testbed, we perform an emulated scalability analysis in Table~\ref{tab:scalability}. We choose Zenoh peer mode as the transport protocol because it is the most reliable solution in decentralized settings, as demonstrated in Table~\ref{tab:communication_quality}. Multiple wired CPU nodes emulate a $6~\si{MB/s}$
per-node bandwidth limit and a simulated interference model while exchanging $200~\si{Hz}$, $128~\si{B}$ control messages. The results show that full $200~\si{Hz}$ throughput is sustained up to $10$ robots, and even at $50$ robots the per-link rate remains above $95$~msgs/s under these assumptions. As team size grows, the dominant bottlenecks are
message size, per-node bandwidth, and per-robot aggregation cost.

\begin{table}[t]
\centering
\caption{Simulated scalability analysis of throughput in a decentralized setup with $200~\si{Hz}$, $128~\si{B}$ control messages under a $6~\si{MB/s}$ per-node bandwidth limit.}
\label{tab:scalability}
\begin{tabular}{@{}lcccc@{}}
\toprule
Number of Robots & 5 & 10 & 30 & 50 \\ \midrule
Throughput per link (msgs/s) & 200.00 & 200.00 & 164.51 & 95.32 \\
\bottomrule
\end{tabular}
\vspace{-5pt}
\end{table}

\vspace{-5pt}
\subsection{Computation Ablation Study}

To guide inference engine selection for different task types, we benchmark the per-module computation cost of collaborative depth perception, control, and task assignment in Table~\ref{tab:computation}, comparing the GPU-based TensorRT engine against the CPU-based ONNX engine.

\begin{table}[!t]
\centering
\caption{Inference engine computation ablation study. All runtimes are reported in milliseconds.}
\label{tab:computation}
\setlength{\tabcolsep}{4pt}
\begin{tabular}{@{}llccc@{}}
\toprule
Module & Inference & Perception & Control & Task \\
& Engine & (ms) & (ms) & Assignment (ms) \\
\midrule
\multirow{2}{*}{Encoder} & ONNX (CPU) & 9.1911 & 16.1 & 0.0691 \\
& TRT (GPU) & 2.4840 & 23.4 & 0.2939 \\
\midrule
\multirow{2}{*}{Aggregation} & ONNX (CPU) & 0.0002 & 16.1  & 0.3851 \\
& TRT (GPU) & 0.0002 & 18.1  & 0.1836 \\
\midrule
\multirow{2}{*}{Decoder} & ONNX (CPU) & 14.8209 & 18.1  & 0.0001 \\
& TRT (GPU) & 2.9565 & 22.4  & 0.0001 \\
\midrule
\multirow{2}{*}{Total Sequential} & ONNX (CPU) & 24.0122 & 50.3 & 0.4532 \\
& TRT (GPU) & 5.4407 & 63.9 & 0.4776 \\
\midrule
\multirow{2}{*}{Total Parallel} & ONNX (CPU) & \textbf{14.8209} & \textbf{18.1} & \textbf{0.3851} \\
& TRT (GPU) & \textbf{2.9565} & \textbf{23.4} & \textbf{0.2939} \\
\bottomrule
\end{tabular}
\vspace{-10pt}
\end{table}

For perception tasks, resource-intensive transformer architectures benefited significantly from GPU parallelization via TensorRT, while their concatenation-based aggregation was less demanding. In contrast, control experiments using small MLPs (up to 64 channels) were faster on CPU with ONNX due to the overhead of CPU-GPU data transfer outweighing any GPU acceleration. Similarly, for task assignment, the simple MLP-based encoder performed better on CPU with ONNX, but the parallel processing capabilities of TensorRT on the GPU were more efficient for the attention mechanism in the aggregation stage, with the identity transform decoder being less resource-intensive.

To validate our parallel pipeline design, we show in Table~\ref{tab:computation} that the parallel pipeline achieves a lower total computation time compared to the sequential pipeline. While the end-to-end latency remains the same, the parallel design allows the system to process new data at a rate limited only by the slowest single stage. This result confirms the benefits of our parallel architecture for real-time applications.
\vspace{-5pt}
\section{Conclusion}~\label{sec:conclusion}
\vspace{-10pt}

In this work, we proposed \mbox{NeuroMesh}, a multi-domain, cross-platform, and modular decentralized framework for multi-robot neural inference. Across perception, control, and decision-making tasks with heterogeneous robots, our experiments demonstrate \mbox{NeuroMesh}'s generality, efficiency, and modularity for real-world deployment.
While our studies focused on learning-based applications, \mbox{NeuroMesh} can also accommodate classical coordination algorithms such as consensus or flocking by reusing its encoding, aggregation, and decoding modules. This highlights its potential as a unified platform for deploying both analytical and learning-based algorithms.
Future works will extend the framework to larger physical teams, broader RF environments, and heterogeneous sensing modalities.
\mbox{NeuroMesh}'s adaptability through custom modules allows it to incorporate community contributions and accommodate future advancements in collaborative robotics research.
\vspace{-5pt}

\bibliographystyle{IEEEtran} 
\bibliography{mybib}
\end{document}